\documentclass[letterpaper, 10 pt, conference]{ieeeconf}  

\IEEEoverridecommandlockouts                              

\usepackage{dsfont}
\usepackage{amsmath}

\usepackage{graphics} 
\usepackage{amsmath} 
\usepackage{amssymb}  
\usepackage{todonotes}

\usepackage{amsfonts}
\usepackage{mathtools}
\usepackage[linesnumbered,algoruled,boxed,vlined, noend]{algorithm2e}
\usepackage{amsthm}
\usepackage{comment}
\usepackage{cite}
\usepackage{wrapfig}
\let\labelindent\relax
\usepackage{enumitem}
\usepackage{overpic}
\usepackage{xspace}

\DeclareMathAlphabet{\mathcal}{OMS}{cmsy}{m}{n} 

\newtheorem{problem}{Problem}

\theoremstyle{definition}

\theoremstyle{remark}

\DeclarePairedDelimiterX{\norm}[1]{\lVert}{\rVert}{#1}

{\end{list}}

\def\trlb{\textsc{TRLB}\xspace}
\def\toro{\texttt{TORO}\xspace}
\def\toroe{\texttt{TORE}\xspace}
\def\toroi{\texttt{TORI}\xspace} 
\usepackage{soul} 
\usepackage[a-2b,mathxmp]{pdfx}[2018/12/22]

\title{\LARGE \bf 
Fast High-Quality Tabletop Rearrangement in Bounded Workspace}

\author{Kai Gao$^{1}$\quad Darren Lau$^{2}$\quad Baichuan Huang$^{1}$\quad  Kostas E. Bekris$^{1}$\quad Jingjin Yu$^{1}$%
\thanks{$^{1}$Department of Computer Science, Rutgers University, NJ, USA. Email: {\tt\small { \{kg627, bh417, kb572, jy512\}}@cs.rutgers.edu}.
}%
\thanks{$^{2}$Department of Computer Science, Cornell University, NY, USA. Email: {\tt\small { \{dl755\}}@cornell.edu}.
}
}

\begin{document}

\maketitle
\thispagestyle{empty}
\pagestyle{empty}


\begin{abstract} 
In this paper, we examine the problem of rearranging many objects on a tabletop in a cluttered setting using overhand grasps. Efficient solutions for the problem, which capture a common task that we solve on a daily basis, are essential in enabling truly intelligent robotic manipulation. In a given instance, objects may need to be placed at temporary positions (``buffers'') to complete the rearrangement,  but allocating these buffer locations can be highly challenging in a cluttered environment. To tackle the challenge, a two-step baseline planner is first developed, which generates a primitive plan based on inherent combinatorial constraints induced by start and goal poses of the objects and then selects buffer locations assisted by the primitive plan. We then employ the ``lazy'' planner in a tree search framework which is further sped up by adapting a novel preprocessing routine. Simulation experiments show our methods can quickly generate high-quality solutions and are more robust in solving large-scale instances than existing state-of-the-art approaches.\\

\noindent source: \href{https://github.com/arc-l/TRLB}{\textcolor{blue}{\texttt{github.com/arc-l/TRLB}}}

\end{abstract}



\section{Introduction}\label{sec:intro}
We study the problem  of  rearranging many objects on a tabletop in a cluttered environment using overhand grasps, where the robot, in a pick-n-place operation, may 
grasp and lift an object, move it around freely, and then place it at a collision free 
pose. This is known as Tabletop Object Rearrangement with Overhand Grasps (\toro), 
which is NP-hard to optimally solve \cite{han2018complexity}.
As a task that humans face all the time, solving \toro autonomously and intelligently is essential in enabling smart robots, at home or in factories.

\begin{figure}[t]
    \centering
    \includegraphics[width=\columnwidth]{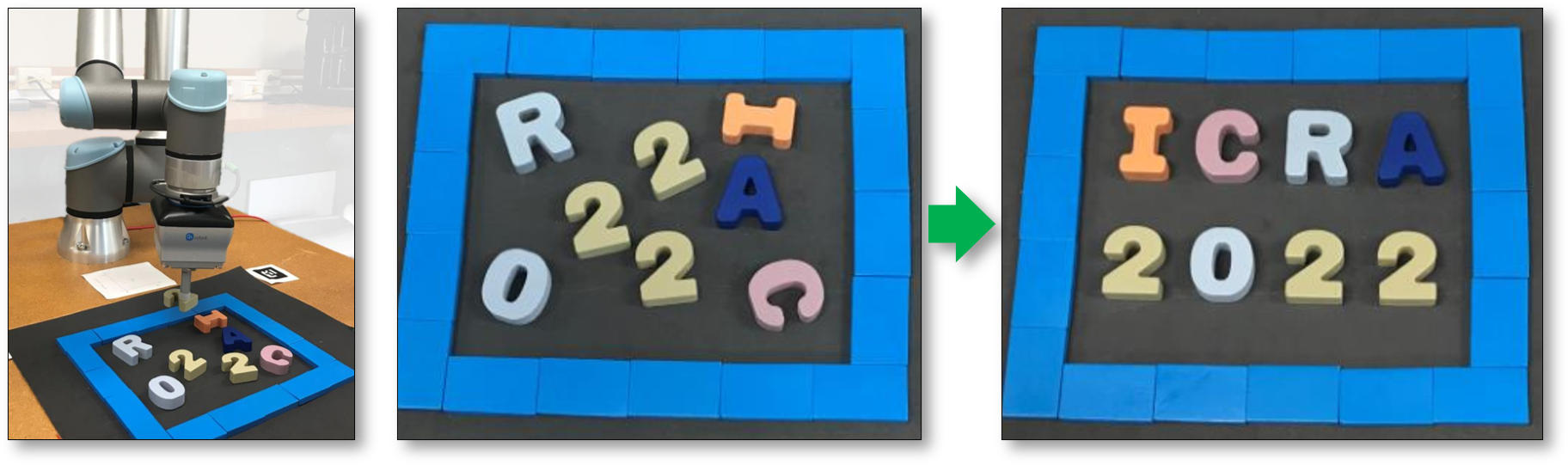}
    \caption{A robot arm rearranges word patterns with overhand grasps, minimizing the number of pick-n-place actions. The letters cannot go out of the bounding box and cannot overlap on the tabletop.}
    \label{fig:ApplicationExample}
\end{figure}

\toro is difficult to optimally solve due to the complex constraints induced by the start and goal poses of the involved objects. 
The combinatorial constraints are generated by (potential) object collisions.
Fig.~\ref{fig:ApplicationExample} shows a robot arm rearranging letters to form ``ICRA 2022'' using a vacuum gripper while having access only to a bounded rectangular area. 
In this example, when letter ``I'' is moved to the goal position, it will collide with ``R'' in its current pose, which results in the \emph{inherent} constraint that ``R'' must move away before ``I'' moves into the goal pose.
%
Such inherent constraints require some objects to be temporarily displaced at ``buffer'' locations, which may induce further \emph{acquired} constraints as there can be many choices of potential buffer locations. Acquired constraints cannot always be avoided, especially when the environment is cluttered. 
Consider again Fig.~\ref{fig:ApplicationExample} for an example. On one hand, if the character ``0'' is transferred to the lower right corner of the workspace, then it needs to be moved before a character ``2'' is moved to the lower right side. On the other hand, if ``0'' is moved to the upper left corner of the workspace, then it needs to be transferred before letter ``I'' can be moved to the goal. 
While the inherent constraints are fully determined by the start and goal configurations, 
acquired constraints depend on the choices made by an algorithm, which significantly increases the size of the search space and the cost of computations.

In tackling \toro, our key contribution is the insight that solutions for untangling inherent constraints \cite{han2018complexity,gaorunning,bereg2006lifting} can assist in resolving the entire problem, and properly using them to develop a fast high-quality solver for dense \toro problems. 
%
We propose \emph{Tabletop Rearrangement with Lazy Buffers} (\trlb), an effective framework for solving \toro in dense/cluttered environments. \trlb first computes a primitive plan only for the inherent constraints and then ``lazily'' selects acquired constraints without the need for a complete feasibility check. 
%
%
Simulation experiments show that \trlb computes high-quality plans for \toro efficiently and is more robust than existing approaches in challenging and practical problems.
Together with the preprocessing routine, \trlb can solve dense 60-object \toro instances in an average of 0.3 seconds, allowing its use in real-time applications. In comparison with state-of-the-art approaches, \trlb computes better solutions two magnitudes faster.

\section{Related Work}
{\bf Rearrangement Approaches:} Object rearrangement is a topic of interest within the broader area of Task and Motion Planning (TAMP). Typical problem definitions in this domain \cite{cosgun2011push, wang2021uniform, gaorunning} involve arranging multiple objects to specific goal positions. Certain problem variations, however, such as NAMO (Navigation among Movable Obstacles) \cite{stilman2005navigation, stilman2007planning, stilman2008planning}, and retrieval problems \cite{ZhangLu-RSS-21, nam2019planning}, 
focus on clearing out a path for a target object or robot. During this process, they are identifying objects that need to be relocated.
Rearrangement may be approached either via simple but inaccurate non-prehensile actions, e.g., pushes \cite{cosgun2011push, king2017unobservable, huang2021dipn}, or more purposeful prehensile actions, such as grasps \cite{krontiris2015dealing, krontiris2016efficiently, wang2021uniform, labbe2020monte}.
Focusing on the inherent combinatorial challenges, some planners use external space for temporary object storage \cite{han2018complexity,gaorunning,bereg2006lifting, nam2019planning}, while others exploit problem linearity to simplify the search  \cite{okada2004environment, stilman2005navigation, stilman2008planning, levihn2013hierarchical}.
By linking rearrangement to established graph-based problems, efficient algorithms have been obtained for various tasks and objectives  \cite{han2018complexity,gaorunning,bereg2006lifting}.
In this paper, we use a plan generated given access to external buffer locations as a ``primitive plan'', which then guides buffer allocation inside the workspace.

{\bf Dependency Graph:} We represent the inherent combinatorial constraints of such rearrangement problems via a dependency graph, which was first applied to general multi-body planning problems \cite{van2009centralized} and then rearrangement \cite{krontiris2015dealing, krontiris2016efficiently}. Choosing different manipulation sequences gives rise to multiple dependency graphs for the same problem instance, which limits the scalability in computing a solution via such representations. 
Prior work \cite{han2018complexity} has applied full dependency graphs to address \toro, showing that the challenge embeds the NP-hard Feedback Vertex Set (FVS) problem and the Traveling Salesperson Problem (TSP).
More recently, some of the authors \cite{gaorunning} examined an optimization objective, \emph{running buffers}, which is the size of the external space needed for the rearrangement task, and also examined an unlabeled setting. Similar graph structures are also used in other robotics problems, such as packing problems \cite{wang2020robot}. 
Deep neural networks have been also applied to detect the embedded dependency graph of objects in a cluttered environment to determine the ordering of object retrieval \cite{ZhangLu-RSS-21}.

{\bf Buffer Identification:} In rearrangement problems, collision-free locations are needed for obstacle displacements. Some previous efforts predetermine buffer candidates and place obstacles to accessible candidate locations when necessary \cite{wang2021uniform, cheong2020relocate}. Others decouple the original problem into successive subproblems \cite{krontiris2016efficiently, wang2021uniform}.
Intuitively speaking, a valid buffer location needs to avoid other objects at their current poses. A backtracking search approach further constrains object displacements given paths of future manipulation actions \cite{stilman2007planning, stilman2008planning}.
Nevertheless, these methods are computationally expensive since they deal with inherent and acquired constraints at the same time.
To reduce the associated computational cost, a lazy strategy can be applied \cite{bohlin2000path, denny2013lazy, hauser2015lazy}, 
which delays path/configuration collision checking. A similar idea is proposed in the current paper. Feasible object locations are obtained with the aid of a ``rough schedule'' of object manipulation actions, 
which is computed given only the inherent constraints. By decoupling inherent and acquired constraints, the proposed method computes high-quality solutions efficiently.

\section{Tabletop Rearrangement\\ with Internal Buffers}
\subsection{Problem Statement}
Consider a 2D bounded workspace $\mathcal W\subset \mathbb R^2$ containing a set of $n$ objects $\mathcal O=\{o_1, ..., o_n\}$. 
Each object is assumed to be an upright \emph{generalized cylinder}. A \emph{feasible arrangement} $\mathcal A =\{p_1, ..., p_n\}$, $p_i = \{x_i, y_i, \theta_i\} \in SE(2)$ is a set of poses for objects in $\mathcal O$, such that 
(1) each object's footprint is contained in $\mathcal W$, and
(2) no two objects collide. 

\begin{figure}[t]
    \centering
    \includegraphics[width=\columnwidth]{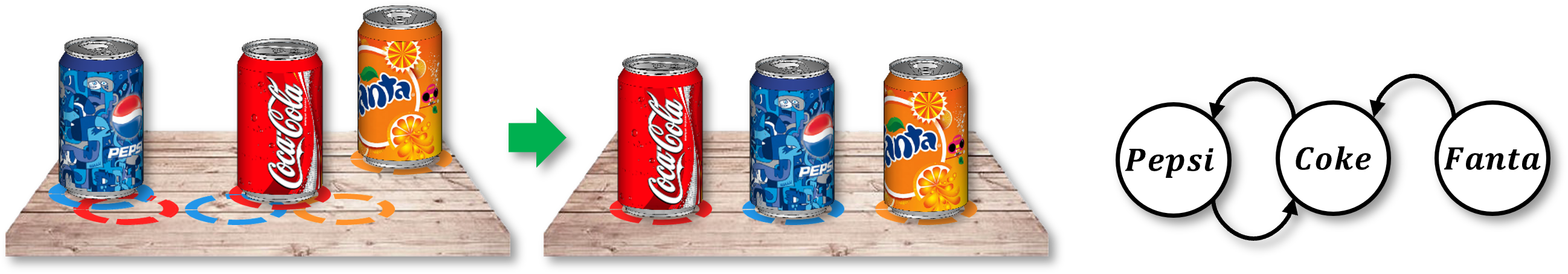}
    \caption{[Left] A rearrangement task involving three soda cans. [Right] The corresponding dependency graph $G$.}
    \label{fig:soda}
\end{figure} 

We consider (overhand) pick-n-place actions to move objects one by one.  A pick-n-place action, represented as an ordered pair $(o, p)$, grasps an object $o$ at its current pose and lifts it above all other objects. It then moves it horizontally, and places it at the target pose $p$ within $\mathcal W$. An action is collision-free if and only if both the current and resulting arrangements are feasible. A plan $P$ from a feasible $\mathcal A_s$ to a feasible $\mathcal A_g$ is a sequence of collision-free pick-n-place actions transforming $\mathcal A_s$ into $\mathcal A_g$. We want to compute feasible plans that minimize the number of pick-n-place actions, which leads to increased system throughput. In summary:

\begin{problem}[\toro w/ Internal Buffers (\toroi)]
Given feasible arrangements $\mathcal A_s=\{p^s_1, ..., p^s_n\}$ and $\mathcal A_g=\{p^g_1, ..., p^g_n\}$, find a feasible plan $P$ sequentially moving objects from $\mathcal A_s$ to $\mathcal A_g$, which minimizes the number of actions.
\end{problem}

\subsection{Dependency Graph and Internal Buffers}

It is not always possible to move an object to its goal pose, which may be occupied by other objects. This leads to \emph{dependencies} between objects, i.e., when the goal pose $p^g_i$ of $o_i$ collides with $o_j$ at its current pose, then $o_i$ depends on $o_j$ and $o_j$ must be moved before moving $o_i$ to its goal.

Dependencies induced by $\mathcal A_s$ and $\mathcal A_g$ give rise to a dependency graph $G$ \cite{van2009centralized, han2018complexity}. 
Fig.~ \ref{fig:soda} provides an example graph. When $G$ is acyclic, the instance is called "monotone" and can be solved with at most $n$ actions moving each object once from its start to its goal pose following the topological order of $G$. Otherwise, some object(s) must be temporarily displaced to break these cycles and complete the task.  We refer to these intermediate poses as "\emph{buffers}", which may be external (i.e., outside $\mathcal W$) or internal (i.e, contained in $\mathcal W$). If the buffers must be internal, the problem is \toroi. Otherwise, the problem is \toro with external buffers, denoted as \toroe. Solving \toroe only requires dealing with inherent constraints defined by $G$. For instance, to solve the problem in Fig.~\ref{fig:soda}, we can move the Pepsi to an external buffer to break the cycle, move the Coke first and then Fanta to their goal locations and finally bring back the Pepsi into the workspace.  With internal buffers, we must find a temporary location for the Pepsi in $\mathcal W$. Due to acquired constraints (as defined in Sec.~\ref{sec:intro}) arising from internal buffer selection, \toroi, the problem we study in this work, is more challenging than \toroe. Intuitively, selecting buffers inside workspace (\toroi) is much more difficult and constrained than using buffers outside the workspace (\toroe) to store displaced objects. 

%

Nevertheless, we show here that plans can be efficiently derived from the minimum \emph{total buffer} solution to \toroe \cite{han2018complexity} and the minimum \emph{running buffer} solution to \toroe \cite{gaorunning}, which computes the minimum number of \emph{concurrent external buffers} needed to solve a \toroe instance. For both objectives, since \toroe has been shown to be computationally intractable \cite{han2018complexity,gaorunning} and is a special case of \toroi, \toroi is also NP-hard. 



%
%

\section{Algorithmic Solutions}
This section first describes a rearrangement solver with lazy buffer allocation (Sec.~\ref{sec:LazyBufferGeneration}), where buffer allocation is delayed after getting a ``rough'' schedule of object movements. To enhance scalability to larger and more cluttered instances, the \trlb framework (Sec.~\ref{sec:PartialPlan}) recovers from buffer allocation failures. Finally, a preprocessing routine helps with further speedups (Sec.~\ref{sec:Preprocess}).

\subsection{Lazy Buffer Allocation}\label{sec:LazyBufferGeneration}

When an object stays at a buffer, it should avoid blocking the upcoming manipulation actions of other objects. Otherwise, either the object in the buffer or the manipulating object has to yield, which increases the number of necessary actions. In other words, we need to carefully choose acquired constraints. If we know the schedule of other objects in advance, a buffer can be selected to minimize unnecessary obstructions. This observation motivates solving the rearrangement problem in two steps: First, compute a \emph{primitive plan}, which is an incomplete schedule ignoring acquired constraints; second, given the incomplete schedule as a reference, generate buffers to optimize the selection of acquired constraints.


\subsubsection{Primitive Plan}
To compute a \emph{primitive plan}, we assume enough free space is available so that no acquired constraints will be created. This transforms the problem into a \toroe problem, where each object is displaced at most once before it moves to the goal pose. Then, an object $o_i \in \mathcal O$ can have three \emph{primitive} actions:   
\begin{enumerate}
    \item $(o_i, s\rightarrow g)$: moving from $p^s_i$ to $p^g_i$;
    \item $(o_i, s\rightarrow b)$: moving from $p^s_i$ to a buffer;
    \item $(o_i, b\rightarrow g)$: moving from a buffer to $p^g_i$.
\end{enumerate}
A primitive plan is a sequence of primitive actions;
computing such a plan is similar to finding a linear vertex ordering \cite{adolphson1973optimal, shiloach1979minimum} of the dependency graph. We use dynamic programming based methods \cite{gaorunning} to achieve this, which minimizes the number of total buffers or running buffers. 

\subsubsection{Buffer Allocation}
Free space inside the workspace $\mathcal{W}$ is scarce in cluttered spaces (e.g.,  Fig.~\ref{fig:density}) and acquired constraints must be dealt with through  the careful allocation of buffers inside $\mathcal{W}$. We apply a greedy strategy to find feasible buffers based on a primitive plan (Algo. \ref{alg:buffer}). The general idea is to incrementally add constraints on the buffers until we find feasible buffers for the whole primitive plan or terminate at a step where there are no feasible buffers for the primitive plan. In Algo.~\ref{alg:buffer},~$\mathcal O_s, \mathcal O_g, \mathcal O_b$ are the sets of objects currently at start poses, goal poses and buffers respectively.

%
%
\begin{algorithm}
\begin{small}
    \SetKwInOut{Input}{Input}
    \SetKwInOut{Output}{Output}
    \SetKwComment{Comment}{\% }{}
    \caption{ Buffer Allocation}
		\label{alg:buffer}
    \SetAlgoLined
		\vspace{0.5mm}
    \Input{$\pi$: a primitive plan; $\mathcal A_s=\{p^s_1,...,p^s_n\}$: start arrangement; $\mathcal A_g=\{p^g_1,...,p^g_n\}$: goal arrangement}
    \Output{$B$: buffers; TerminatingStep: the action step where buffer generation fails, $\infty$ if Success.}
		\vspace{0.5mm}
		$\mathcal O_s$ $\leftarrow$ $\mathcal O$; 
		$\mathcal O_g$, $\mathcal O_b$ $\leftarrow$ $\emptyset$; 
		$B \leftarrow$ RandomPoses($\mathcal O$)\\
		\For{   $(o_i, m)\in \pi$}{
    		\If{$m$ is s $\rightarrow$ b}{
        		$\mathcal O_b$.add($o_i$)\\
        		Constraints[$o_i$]$\leftarrow$GetPoses($\mathcal O_s \bigcup \mathcal O_g - \{o_i\}$)\\
    		}
    		\ElseIf{$m$ is b $\rightarrow$ g}{
        		\lFor{$o\in \mathcal O_b\backslash\{o_i\}$}{
        		Constraints[$o$].add($p^g_i$)
    		}
    		}
    		\Else{
        		\lFor{$o\in \mathcal O_b$}{
        		    Constraints[$o$].add($p^g_i$)
        		}
    		}
    		Success, $B'$ $\leftarrow$ BufferGeneration($\mathcal O_b$, Constraints, $B$)\\
    		\If{Success}{
    		$B$ $\leftarrow$ $B'$\\
    		$\mathcal O_s, \mathcal O_g, \mathcal O_b \leftarrow$ UpdateState($\mathcal O_s, \mathcal O_g, \mathcal O_b$)\\
    		}
    		\lElse{
    		\Return $B$, $\pi$.index(action)
    		}
		}
		\vspace{0.5mm}
		\Return $B$, $\infty$\\
\end{small}
\end{algorithm}

We start with $\mathcal A_s$ where all the objects are at start poses and the buffers are initialized at random poses (line 1). Each action in $\pi$ indicates an object $o_i$ that is manipulated and the action $m$ performed (line 2). If $o_i$ is moved to a buffer (line 3),  then we add it into $\mathcal O_b$ (line 4). The current poses of other objects in $O_s\bigcup O_g$ are seen as fixed obstacles for $o_i$ (line 5). If $o_i$ is leaving the buffer (line 6), then other objects in $\mathcal O_b$ should avoid the goal pose $p^g_i$ of $o_i$ (line 7). If $o_i$ is moving directly from $p^s_i$ to $p^g_i$ (line 8, the ``else'' corresponds to $m$ being $s\to g$, e.g., directly go from start to goal), then all buffers for objects in the current $\mathcal O_b$ need to avoid $p^g_i$ (line 9). After setting up acquired constraints, we generate new buffers for objects in $O_b$ to satisfy these constraints by either sampling or solving an optimization problem (line 10). Old buffers in $B$ satisfying new constraints will be directly adopted. If feasible buffers are found (line 11), then buffers and object states will be updated (line 12-13). Otherwise, we return the feasible buffers computed and record the terminating step  of the algorithm (line 14). In the case of a failure, the returned buffers provide a \emph{partial plan}.

\begin{figure}
    \vspace{2mm}
    \centering
    \includegraphics[width=\columnwidth]{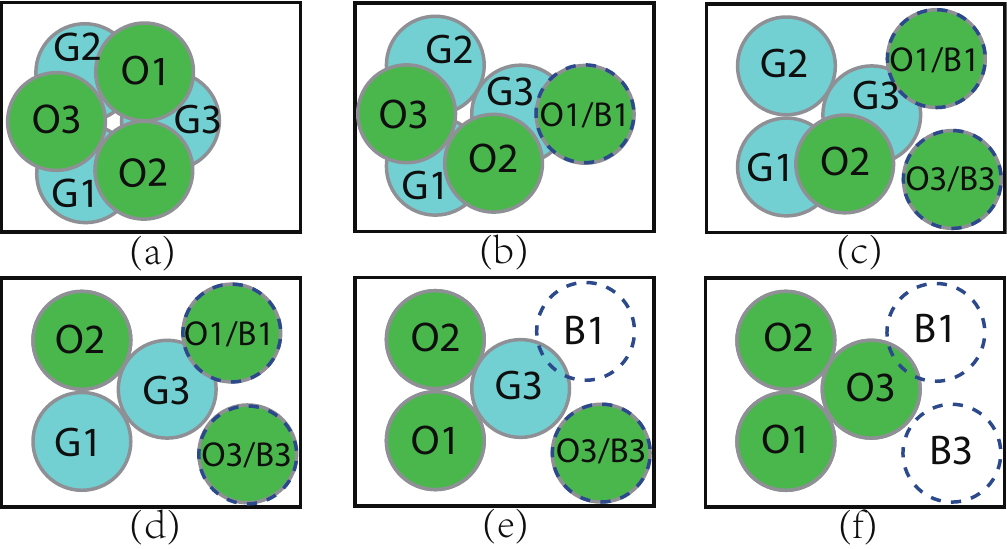}
    \vspace{-.25in}
    \caption{A working example with three objects defined in (a).
    The primitive plan is [($o_1$, s $\rightarrow$ b), ($o_3$, s $\rightarrow$ b), ($o_2$, s $\rightarrow$ g), ($o_1$, b $\rightarrow$ g), ($o_3$, b $\rightarrow$ g)].  Figures (b)-(f) show the steps of Alg. 1 after each action. The transparent discs with dashed line boundaries ($B_i$) represent the buffers satisfying constraints up to each step. For each object $o_i$, the green ($O_i$) and cyan ($G_i$) discs represent the current and goal poses respectively.}
    \label{fig:AlgoExample}
\end{figure}

Fig.~\ref{fig:AlgoExample} illustrates the buffer allocation process via an example. The green, cyan, and transparent discs represent the current poses, goal poses and allocated buffers respectively. When we move $o_1$ to a buffer $B_1$ (Fig. \ref{fig:AlgoExample}(b)),  it only needs to avoid collision with $p^s_2$ and $p^s_3$.  But as we move $o_3$ to a buffer,  $B_1$ needs to avoid $o_3$'s buffer $B_3$ as well. To satisfy the added constraint, $B_1$ will be reallocated. Since the new buffers $B_1$ and $B_3$ (Fig.~\ref{fig:AlgoExample}(c)) satisfy the constraints added in the following steps, they need not to be relocated. Note that the buffer originally selected for $o_1$ but then replaced will not appear in the resulting plan, i.e., $o_1$ will move directly to the new buffer (Fig.~\ref{fig:AlgoExample}(c)-(f)). Algo. \ref{alg:buffer} works with one strongly connected component of the dependency graph at a time, treating objects in other components as fixed obstacles.


Once the feasible buffers are found, all the primitive actions can be transformed into feasible pick-n-place actions inside the workspace. And therefore, the primitive plan can be transformed into a rearrangement plan moving objects from $\mathcal A_s$ to $\mathcal A_g$. The function BufferGeneration is implemented by either sampling or solving an optimization problem, both of which are discussed below.

\paragraph{Sampling}\label{sec:Sampling}
Given the object poses that buffers need to avoid so far, feasible buffers can be generated by sampling poses inside the free space. When objects stay in buffers at the same time,we sample buffers one by one and previously sampled buffers will be seen as obstacles for latter ones.

\paragraph{Optimization}\label{sec:Optimization}
For cylindrical objects $o_i$ at $(x_i,y_i)$ with radius $r_i$, and $o_j$ at $(x_j,y_j)$ with radius $r_j$, they are collision-free when
$(x_i-x_j)^2+(y_i-y_j)^2 \geq (r_i+r_j)^2$ holds. By further restricting the range of buffer centroids to assure they are in the workspace, the buffer allocation problem can be transformed into a quadratic optimization problem. For objects with general shapes, collision avoidance cannot be presented by inequalities of object centroids. We can construct the optimization problem with $\phi$ functions of the objects \cite{chernov2010mathematical} and solve the problem with gradients.

\subsection{Tabletop Rearrangement with Lazy Buffers (\trlb)}\label{sec:PartialPlan}
The \trlb framework builds on the insight that a new \toroi instance is generated when lazy buffer allocation fails. The new instance has the same goal $\mathcal A_g$ as the original one but some progress has been made in solving the \toroi task. There are two straightforward implementations of \trlb: forward search and bidirectional search. In the first case, by accepting partial solutions, a rearrangement plan can be computed by developing a search tree $T$ rooted at $\mathcal A_s$.  In the search tree $T$, nodes are feasible arrangements and edges are partial plans containing a sequence of collision-free actions. When buffer allocation fails, we add the resulting arrangement into the tree and resume the rearrangement task from a random node in $T$.  This randomness and the randomness in primitive plan computation and buffer allocation allows \trlb to recover from failures. 
\vspace{0.05in}


\begin{algorithm}[h]\label{alg:BST}
\begin{small}
    \SetKwInOut{Input}{Input}
    \SetKwInOut{Output}{Output}
    \SetKwComment{Comment}{\% }{}
    \caption{\trlb with Bidirectional Search}
		\label{alg:BS}
    \SetAlgoLined
		\vspace{0.5mm}
    \Input{$\mathcal A_s$, $\mathcal A_g$, $max\_time$}
    \Output{ Search trees: $T_1$, $T_2$}
		\vspace{0.5mm}
		$T_1$.root, $T_2$.root$\leftarrow \mathcal A_s, \mathcal A_g$\\
		\While{not exceeding $max\_time$}{
		$\mathcal A_{rand}\leftarrow$ RandomNode($T_1$)\\
		$\mathcal A_{new1} \leftarrow$ LazyBufferAllocation($\mathcal A_{rand}$, $T_2$.root)\\
		$T_1$.add($\mathcal A_{new1}$)\\
		\lIf{$\mathcal A_{new1}$ is $T_2$.root}{\Return $T_1$, $T_2$}
		$\mathcal A_{near}\leftarrow$ NearestNode($\mathcal A_{new1}$, $T_2$)\\
		$\mathcal A_{new2} \leftarrow$ LazyBufferAllocation($\mathcal A_{near}$, $\mathcal A_{new1}$)\\
		$T_2$.add($\mathcal A_{new2}$)\\
		\lIf{$\mathcal A_{new2}$ is $\mathcal A_{new1}$}{\Return $T_1$, $T_2$}
		$T_1, T_2 \leftarrow T_2, T_1$\\
		}
\end{small}
\end{algorithm}

In bidirectional search, two search trees rooted at $\mathcal A_s$ and $\mathcal A_g$ are developed. This more involved procedure is shown in Algo. \ref{alg:BS}, which computes two search trees that connect $\mathcal A_s$ and $\mathcal A_g$.  In line 1, the trees are initialized. For each iteration, we first rearrange between a random node $\mathcal A_{rand}$ on $T_1$ to the root node of $T_2$ (line 3-5). The function LazyBufferAllocation refers to the overall algorithm developed in Sec. \ref{sec:LazyBufferGeneration}. A found path yields a feasible plan for \toroi (line 6). Otherwise, we rearrange between the new arrangement $\mathcal A_{new1}$ and its nearest neighbor in $T_2$ (line 7-9). If a path is found, then we find a feasible rearrangement plan for \toroi (line 10). Otherwise, we switch the trees and attempt rearrangement from the opposite side (line 11).

\subsection{Preprocessing}\label{sec:Preprocess}
\begin{figure}[h]
    \centering
    \includegraphics[width=\columnwidth]{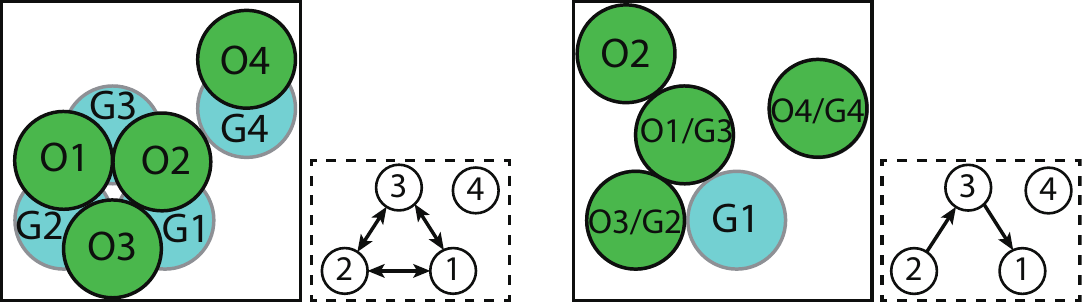}
    \vspace{-.2in}
    \caption{A four-object example of preprocessing. The green and cyan discs represent current and goal arrangements respectively. Before preprocessing (left), two buffers need to be allocated synchronously. After  preprocessing (right), the problem becomes monotone.}
    \label{fig:preprocessing}
    \vspace{-.1in}
\end{figure}

In dense environments, allocating buffers is hard, motivating minimizing the number of running buffers \cite{gaorunning}, which is generally low even in high density settings if we treat objects as \emph{unlabeled}. Based on this, for each component of the dependency graph that is not a single vertex/cycle, we reduce the running buffer size to at most 1 by first solving an \emph{unlabeled} instance \cite{gaorunning}. After preprocessing, we obtain a \toroi requiring at most one running buffer. Fig.~\ref{fig:preprocessing} shows an example of preprocessing.  $o_1$, $o_2$ and $o_3$ form a complete graph, where at least two objects need to be placed at buffers simultaneously. We conduct preprocessing of the three-vertex component by moving $o_2$ to a buffer position, $o_1$ to $p^g_3$ and $o_3$ to $p^g_2$. $o_2$ will not move to $p^g_1$ since it does not occupy other goal poses. The preprocessing step needs one buffer and the resulting rearrangement problem is monotone.

%



\section{Experiments}
We implemented the algorithms of the \trlb framework in Python. Simulated experiments use environments with different density levels $\rho$, defined as the proportion of the tabletop surface occupied by objects, 
i.e., $\rho:=(\Sigma_{o_i\in \mathcal O} S_{o_i})/S_{\mathcal W}$, 
where $S_{o_i}$ is the base area of $o_i$ and $S_{\mathcal W}$ is the area of $\mathcal W$. 

\begin{figure}[h!]
\centering
\includegraphics[width=0.225\columnwidth]{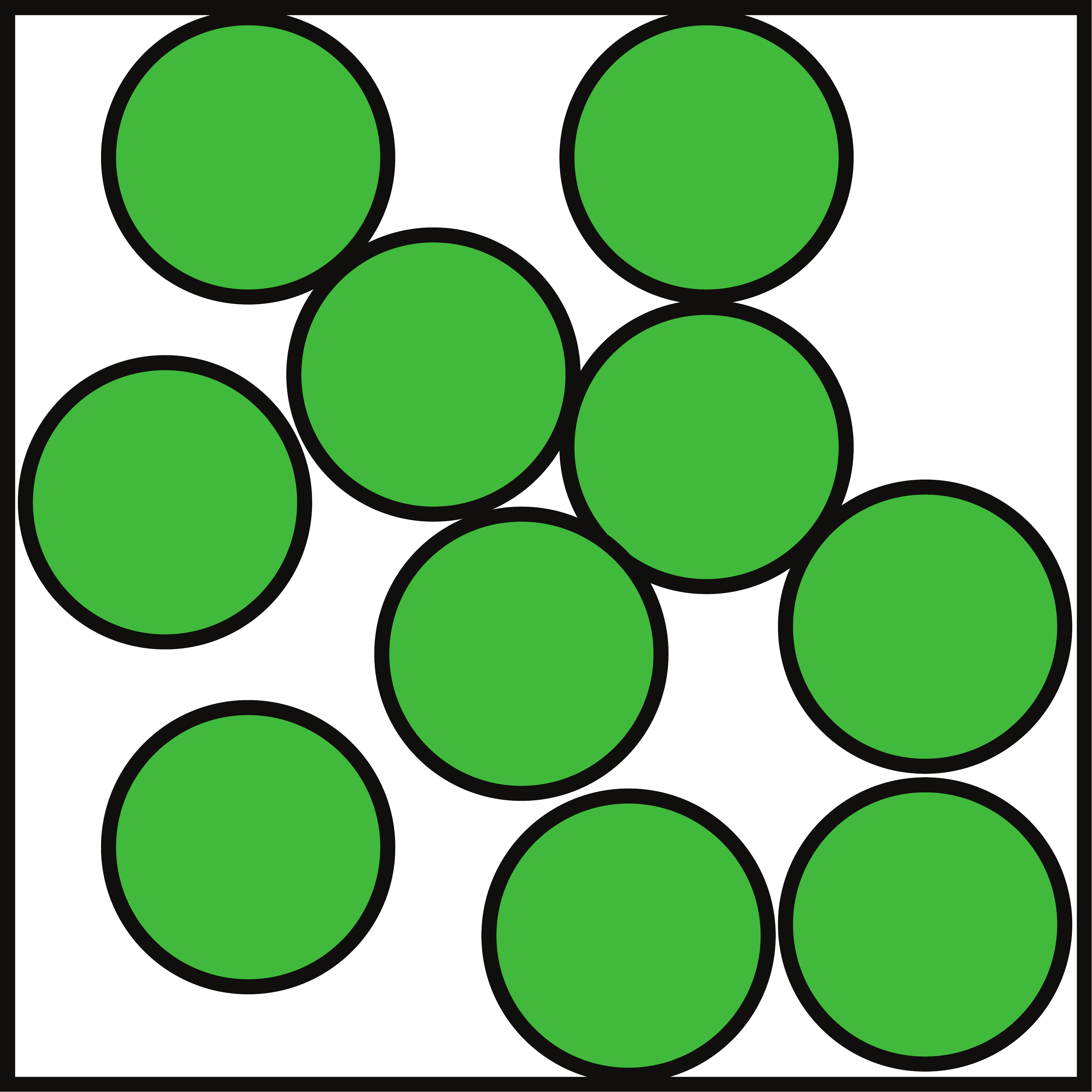}\hspace{10mm}
\includegraphics[width=0.225\columnwidth]{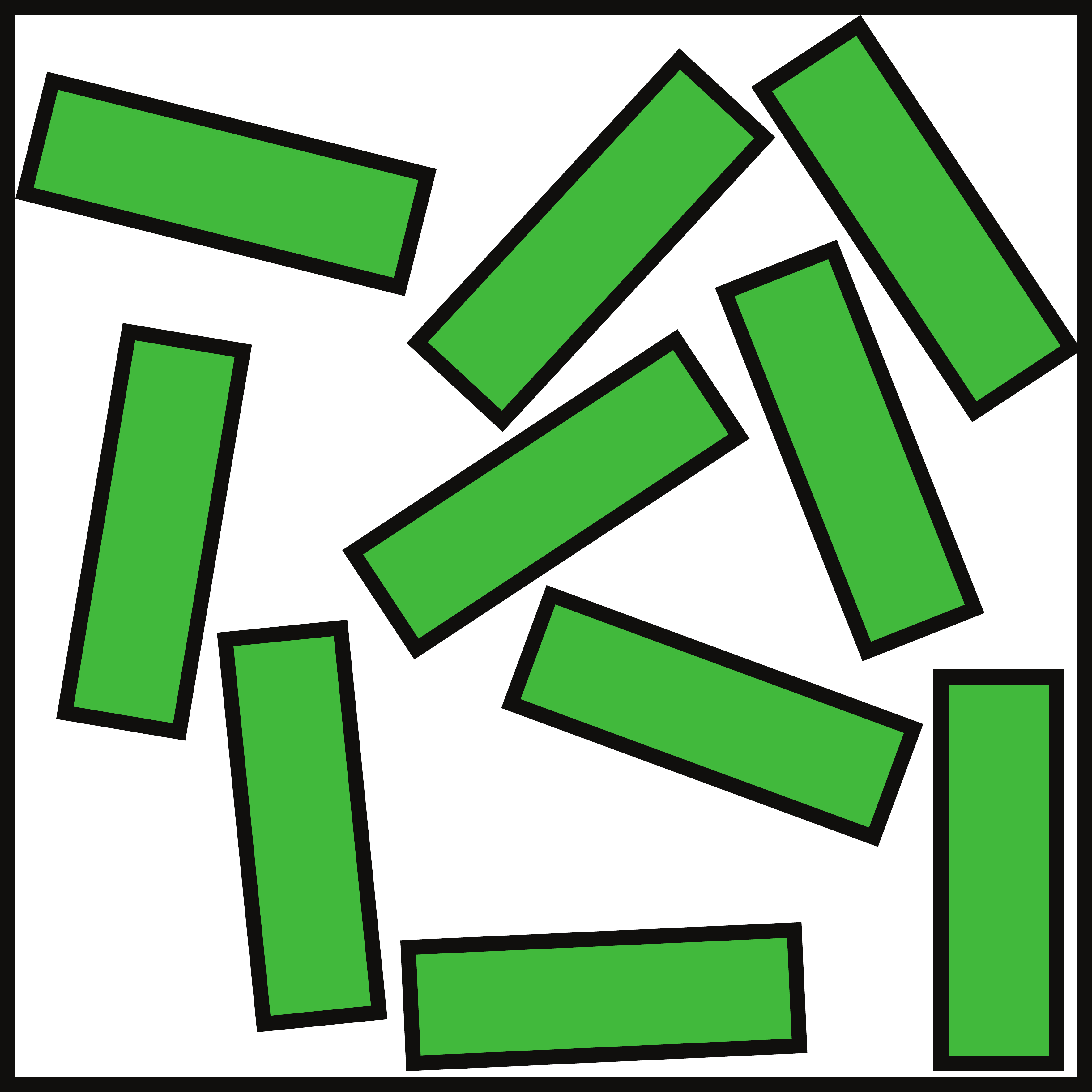}
\vspace{-.05in}
    \caption{[left] 10 cylinders with $\rho=0.5$, [right] 10 cuboids with $\rho=0.4$.}
    \label{fig:density}
\end{figure}

Fig.~\ref{fig:density} shows a dense 10-cylinder arrangement with $\rho=0.5$, and a dense 10-cube arrangement with $\rho=0.4$. The experiments are executed on an Intel$^\circledR$ Xeon$^\circledR$ CPU at 3.00GHz. Each data point is the average of $30$ test cases except for unfinished trials, if any, given a time limit of $300$ sec. per test case.

\subsection{Ablation Study for Cylindrical Objects}
We first present experiments with cylindrical objects to compare lazy buffer generation algorithms given different options, including:
(1) Primitive plan computation: running buffer minimization (RBM), total buffer minimization (TBM), random order (RO); (2) Buffer allocation methods: optimization (OPT), sampling (SP); (3) High level planners: one-shot (OS), forward search tree (ST), bidirectional search tree (BST); and (4) With or without preprocessing (PP).
%
Here, the one-shot (OS) planner is using primitive plans and buffer allocation (Sec.~\ref{sec:LazyBufferGeneration}) without tree search (Sec.~\ref{sec:PartialPlan}). In OS, we attempt to compute a feasible rearrangement plan up to $30|\mathcal O|$ times before announcing a failure.
Notice that at least $|\mathcal O|$ actions are required for solving any instance. 

A full \trlb algorithm is a combination of components, e.g.,  RBM-SP-BST stands for using the primitive plans that minimize running buffer size,  performing buffer allocation by sampling, maintaining a bidirectional search tree, and doing so without preprocessing.

\begin{figure}[h!]
    \vspace{2mm}
\centering
    \begin{overpic}[width=1\columnwidth]{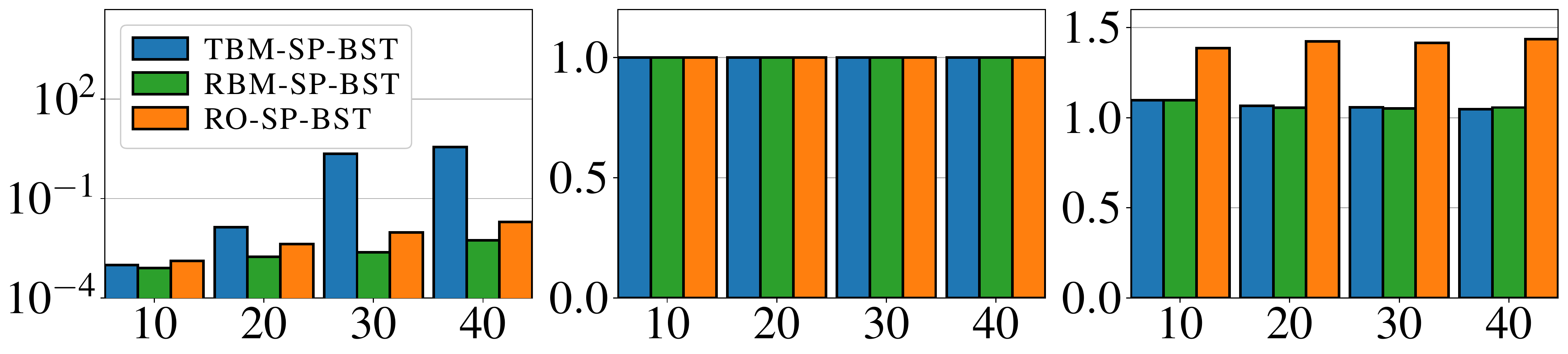}
    \put(-1.5,-22.5){ \includegraphics[width=1\columnwidth]{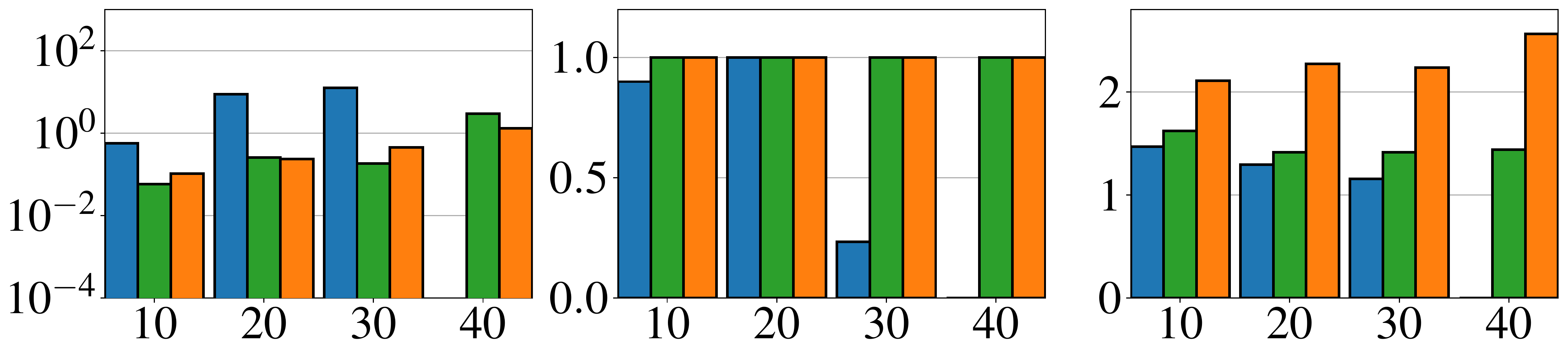}}
    \end{overpic}
    \vspace{13mm}
    \caption{Comparison of primitive planners with $10$-$40$ cylinders and density levels $\rho=0.3$ (top), 0.5 (bottom) (left: computation time in seconds; middle: success rate; right: number of actions as multiples of $|\mathcal O|$).}
    \label{fig:Primitive}
\end{figure}

For evaluation, we first compare the primitive plan computation options, using sampling-based buffer allocation, bidirectional tree search and no preprocessing.  TBM and RBM plans are computed using dynamic programming based solvers \cite{gaorunning}. The results are shown in Fig.\ref{fig:Primitive}. 
Even though plans generated by TBM-SP-BST are slightly shorter than RBM-SP-BST, TBM-SP-BST is less scalable as either the density level or the number of objects in the workspace increases. Compared to RBM plans, individual RO plans can be generated almost instantaneously but we don't see much benefit in computation time for the overall algorithm. 
The results indicate that RBM should be used for primitive plan computation as it results in efficient and high-quality solutions.

\begin{figure}[h!]
    \vspace{2mm}
\centering
    \begin{overpic}[width=1\columnwidth]{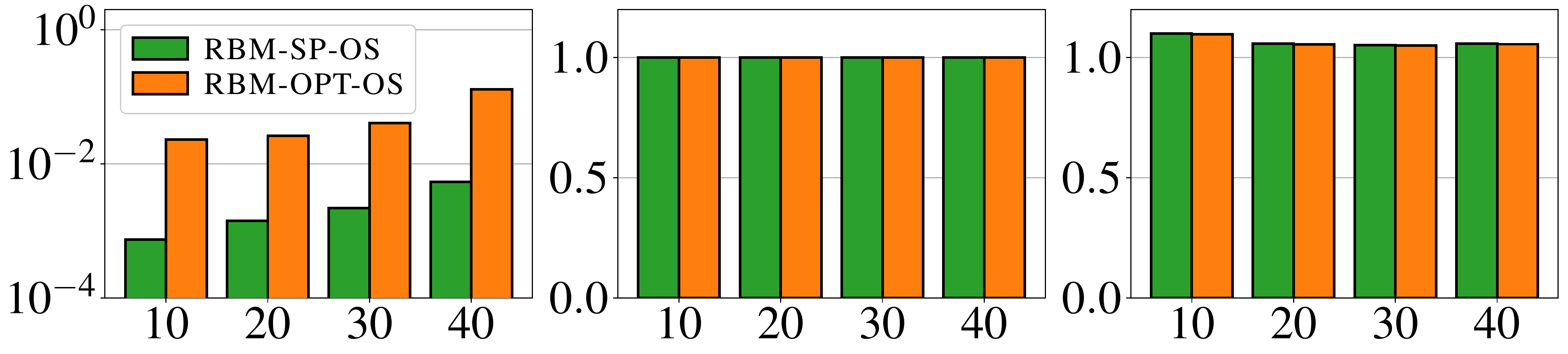}
    \put(-1.5,-22.5){ \includegraphics[width=1\columnwidth]{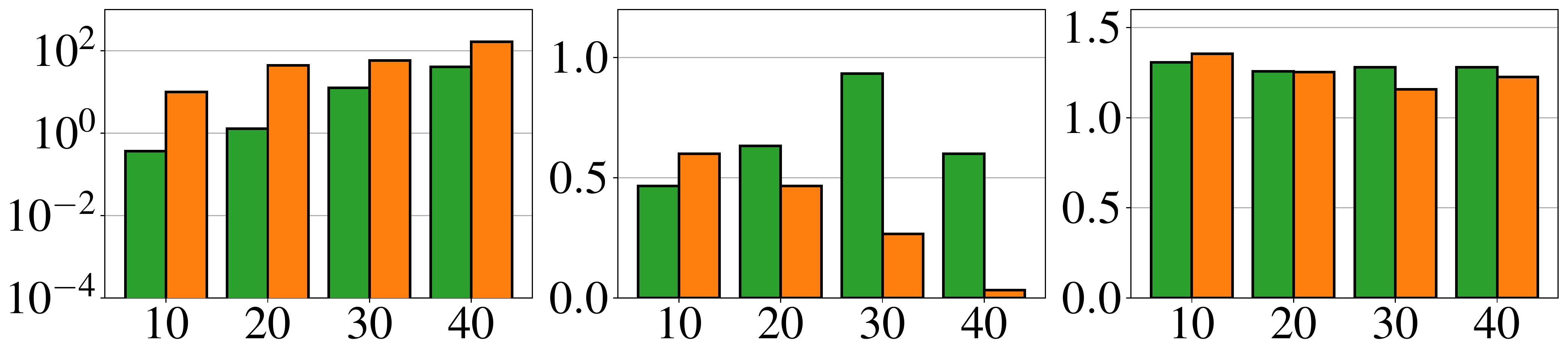}}
    \end{overpic}
    \vspace{13mm}
    \caption{Comparison of buffer allocation methods with $10$-$40$ cylinders at density levels $\rho=0.3$ (top), 0.5 (bottom) (left: computation time in seconds; middle: success rate; right: number of actions as multiples of $|\mathcal O|$).}
    \label{fig:BufferAllocation}
\end{figure}

In Fig. \ref{fig:BufferAllocation}, buffer allocation methods are compared using the RBM primitive planner and the OS high-level planner. Optimization-based allocation guarantees completeness and generates high-quality plans but it is computationally expensive. When $\rho=0.5$, the success rate tends to be low in instances with a small number of objects.  That is because for the given density level, the smaller the 
number of objects, the larger the object size relative to the environment, and the smaller the configuration space size relative to the environment. Thus, precisely allocating buffer locations with OPT is helpful in these cases.

\begin{figure}[h!]
    \vspace{1mm}
\centering
    \begin{overpic}[width=1\columnwidth]{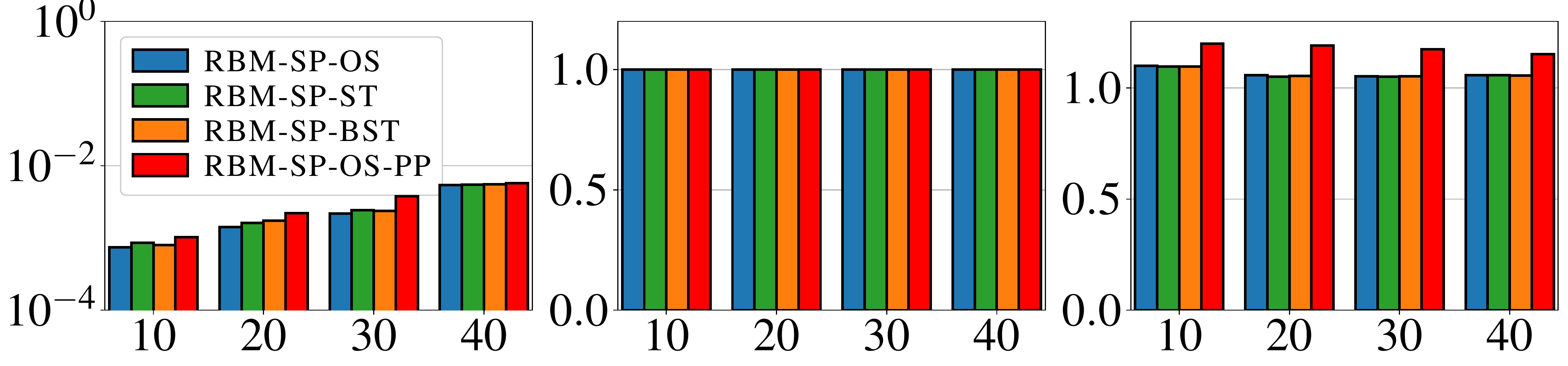}
    \put(-1.5,-22.5){ \includegraphics[width=1\columnwidth]{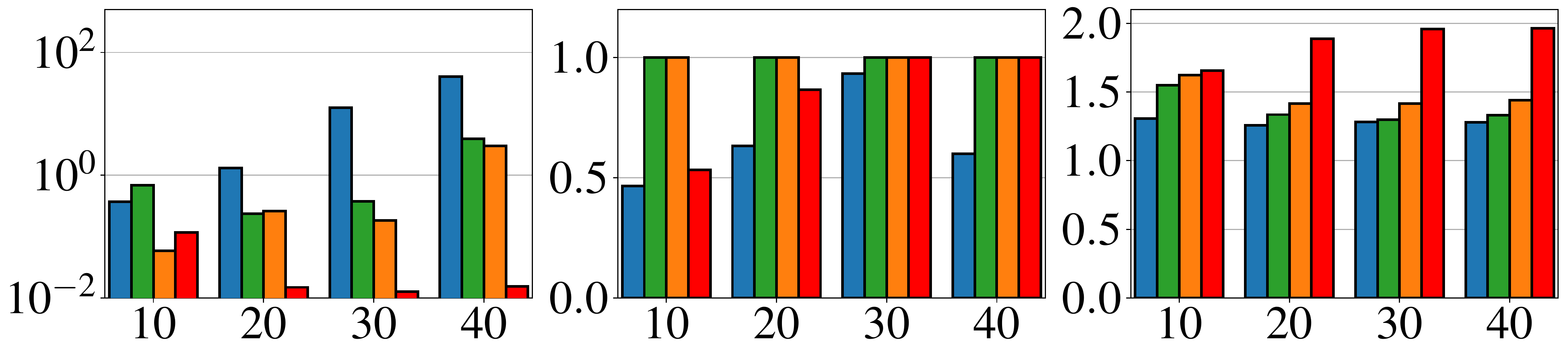}}
    \end{overpic}
    \vspace{13mm}
    \caption{Comparison among high-level frameworks and OS with preprocessing. There are 10-40 cylinders in the workspace at density levels $\rho=0.3$ (top), $0.5$ (bottom) (left: computation time in seconds; middle: success rate; right: number of actions as multiples of $|\mathcal O|$).}
    \label{fig:HighLevel}
\end{figure}

The effectiveness of the high-level planners and preprocessing are shown in Fig. \ref{fig:HighLevel}, which suggests that ST, BST and preprocessing are all effective in increasing success rate in dense environments. In addition, preprocessing significantly speeds up computation in large scale dense cases at the price of extra actions to execute preprocessing. By simplifying the dependency graph with preprocessing, less time is needed to compute a primitive plan.

\begin{figure}[h!]
    \vspace{2mm}
\centering
\includegraphics[width=0.6\columnwidth]{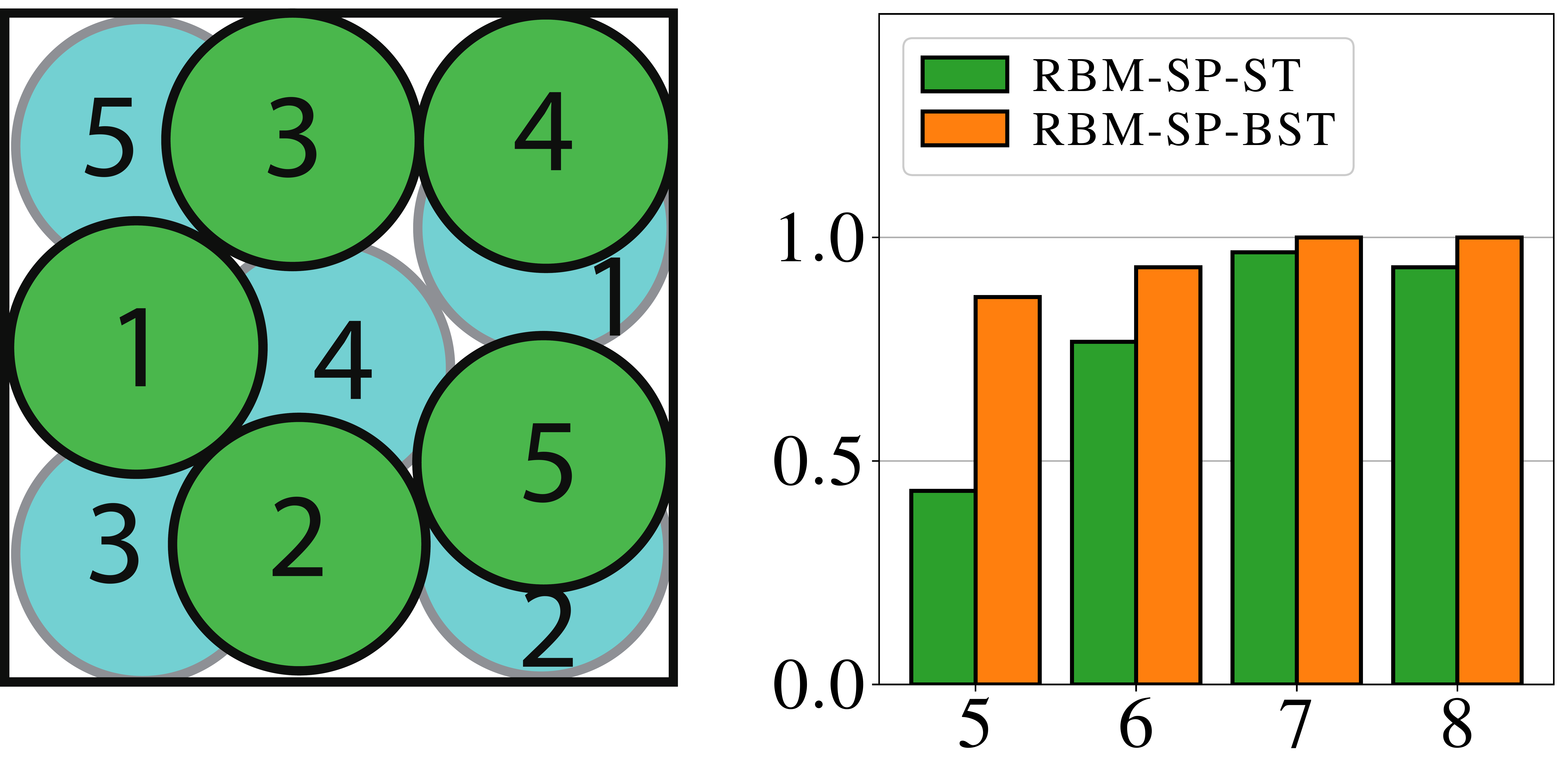}
\vspace{-0.1in}
    \caption{Comparison between ST and BST frameworks with ``dense-small'' instances where 5-8 cylinders packed in the environment and density level $\rho=0.5$. [left] An example with 5 cylinders, [right] success rate of methods.}
    \label{fig:HighLevelDenseSmall}
\end{figure}

The robustness of ST and BST are further evaluated with ``dense-small'' instances where a few objects are packed densely (Fig. \ref{fig:HighLevelDenseSmall}). The bidirectional search tree has a higher success rate in these cases, especially in 5-object instances.

In addition to the above evaluations, we also tried integrating the preprocessing into the BST framework (RBM-SP-BST-PP), which speeds up the computation:  for $60$-object instances with $\rho=0.5$,  only $63\%$ of them can be solved by RBM-SP-BST in 300 seconds. All of them, however, can be solved together with the preprocessing and the solution time averaged 0.29 seconds. Similarly to the results in Fig.~\ref{fig:HighLevel}, preprocessing makes the solution plan much longer than necessary (needs around $30\%$ more actions than RBM-SP-BST). Based on the analysis on computation time, success rate, and solution quality, RBM-SP-BST is the best overall combination, and preprocessing significantly speeds up the solver with a reduction of solution quality.

\subsection{Comparison with Alternatives for Cylindrical Objects}
\begin{figure}[h!]
    \vspace{2mm}
\centering
    \begin{overpic}[width=1\columnwidth]{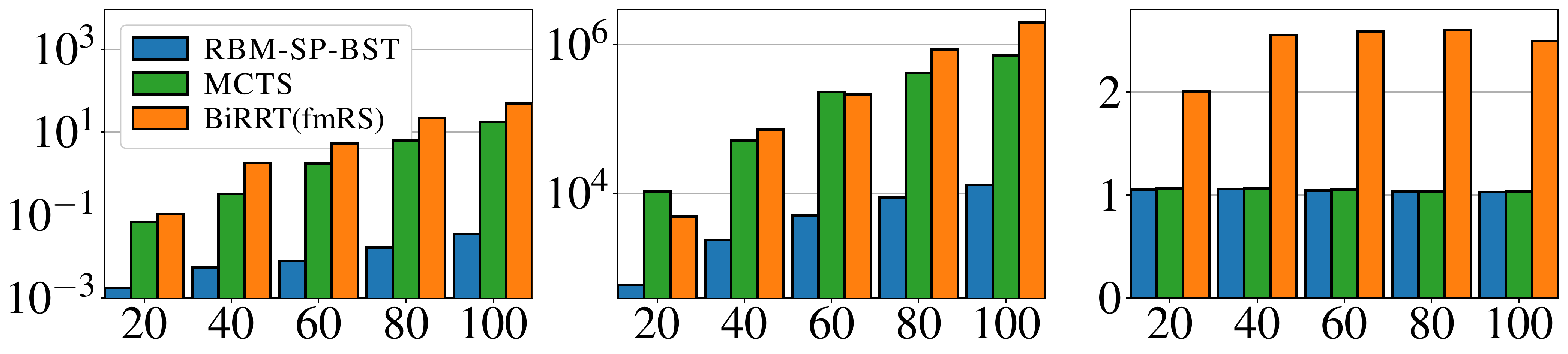}
    \end{overpic}
    \vspace{-.1in}
    \caption{Comparison of algorithms with 20-100 cylinders at density level $\rho=0.3$ (left: computation time in seconds; middle: number of collision checks; right: number of actions as multiples of $|\mathcal O|$).}
    \label{fig:LargeScale}
\end{figure}

We compare the proposed method RBM-SP-BST with BiRRT(fmRS) \cite{krontiris2016efficiently} and a MCTS planner \cite{labbe2020monte}, which, to the best of our knowledge, are state-of-the-art planners for \toroi. The MCTS planner is a C++ solver, while the other two methods are implemented in Python. Besides success rate, solution quality, and computation time, 
we also compare the number of collision checks which are time-consuming in most planning tasks.  In Fig.~\ref{fig:LargeScale}, we compare the methods in large scale problems with $\rho=0.3$. The success rate is $100\%$ for all.
Our method, RBM-SP-BST, avoids repeated collision checks due to the use of the dependency graph. BiRRT(fmRS), which only uses dependency graphs locally, spends a lot of time and conducts a lot of collision checks to generate random arrangements. MCTS generates solutions with similar optimality but does so also with a lot of collision checking, which slows down computation.  We note that a value of $1$ in the right figure (number of actions) is the minimum possible, so both RBM-SP-BST and MCTS compute high-quality solutions, which RBM-SP-BST does slightly better. To sum up, in sparse large scale instances, RBM-SP-BST is two magnitudes faster and conducts much fewer collision checks than the alternatives.

\begin{figure}[h!]
    \vspace{2mm}
\centering
    \includegraphics[width=1\columnwidth]{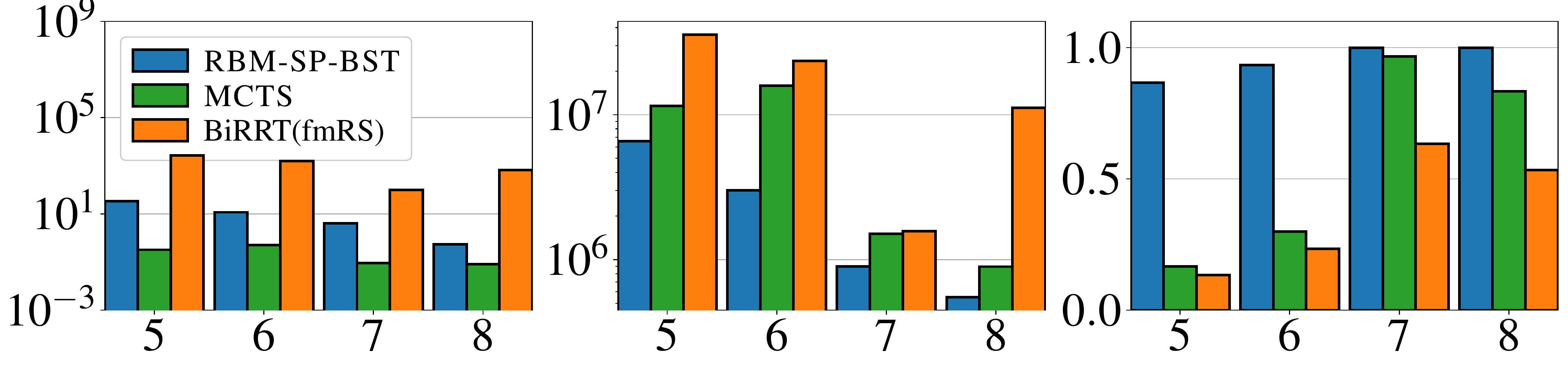}
    \caption{Comparison of methods on ``dense-small'' instances where 5-8 objects are packed in an environment with $\rho=0.5$ (left: computation time in seconds; middle: number of collisions; right: success rate).}
    \label{fig:DenseSmallComaprison}
\end{figure}

Next, we compare the methods in ``dense-small'' instances (Fig.~\ref{fig:DenseSmallComaprison}). Here, RBM-SP-BST is the only method that maintains high success rate in these difficult cases.

\begin{figure}[h!]
    \vspace{2mm}
\centering
\includegraphics[width=0.3\textwidth]{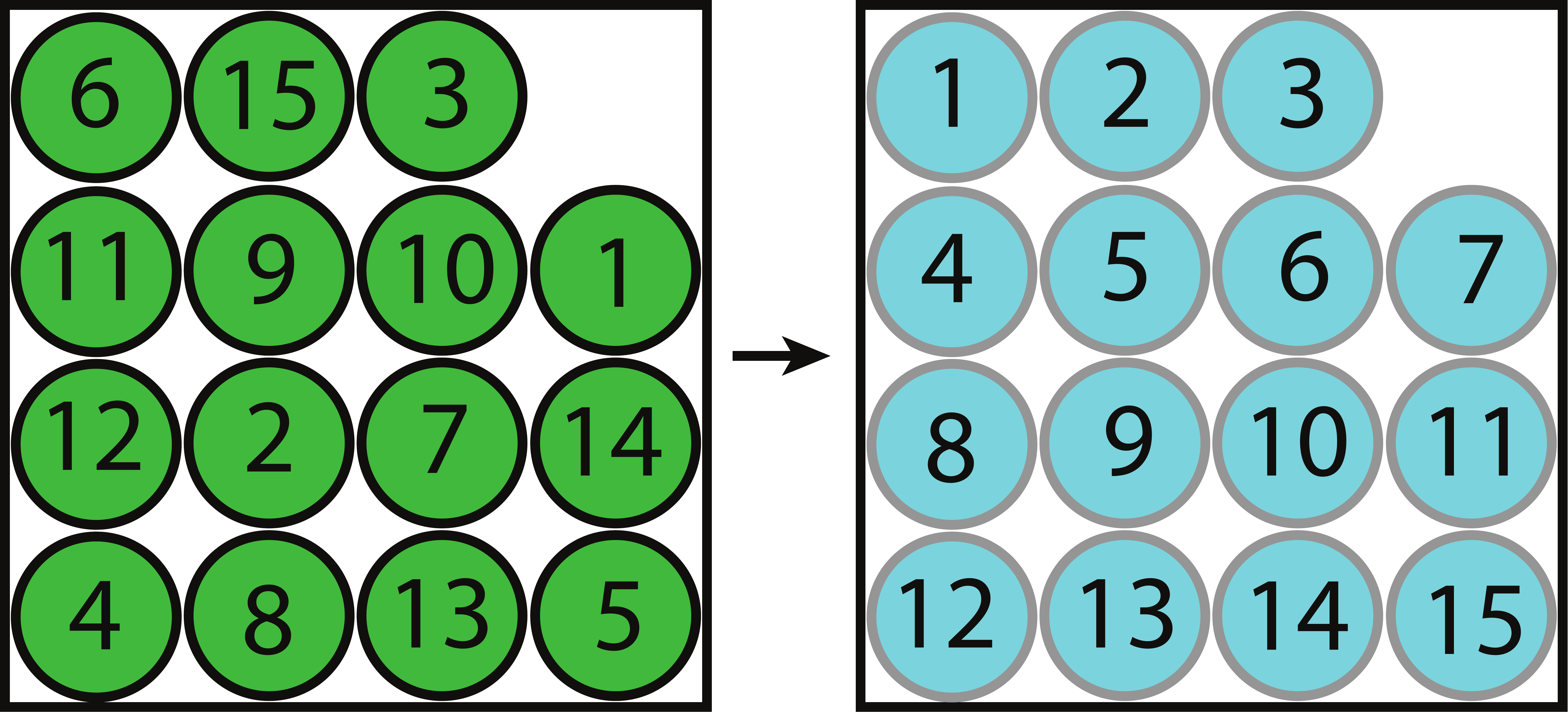}\hspace{2mm}
    \includegraphics[width=0.3\columnwidth]{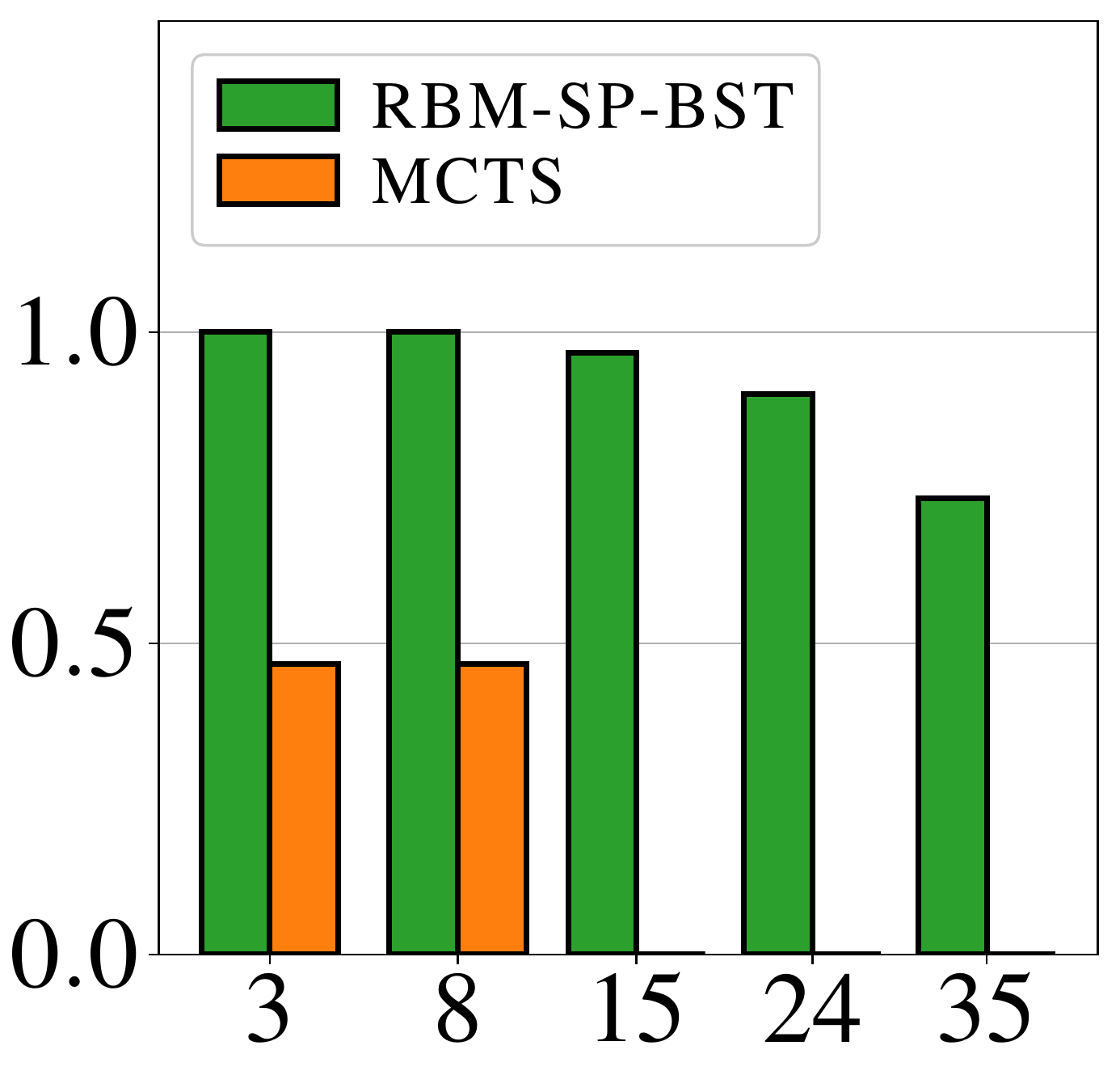}
    \caption{Comparison among methods in lattice instances with 3-35 objects. [left] lattice example; [right] success rate}
    \label{fig:lattice}
\end{figure}

We further compare the performance of RBM-SP-BST and MCTS in lattice rearrangement problems, which are recently studied in the literature \cite{yurearrangement}. 
An example with 15 objects is shown in Fig.~\ref{fig:lattice}[left]. 
In the start and goal arrangements, 
gaps between adjacent objects are set to be 0.01 object radius,
and thus buffer allocation is challenging for sampling-based methods. 
While MCTS tries all the actions on each node, RBM-SP-BST is able to detect the embedded combinatorial object relationship via the dependency graph and therefore needs less buffer allocation calls. As shown in Fig.~\ref{fig:lattice}[right], RBM-SP-BST has much higher success rate in lattice rearrangement tasks.

\subsection{Cuboid Objects}
Because the MCTS solver only supports cylindrical objects, we only compare RBM-SP-BST and BiRRT(fmRS) in the cuboids setup (Fig.~\ref{fig:density}[right]).
When $\rho=0.3$, RBM-SP-BST computes high quality solutions efficiently, 
while BiRRT(fmRS) can only solve instances with up to 20 cuboids.
We mention that, when $\rho=0.4$, BiRRT(fmRS) cannot solve any instance, 
but RBM-SP-BST can solve 50-object rearrangement problems in 28.6 secs on average.

\begin{figure}[h!]
    \vspace{2mm}
\centering
    \includegraphics[width=1\columnwidth]{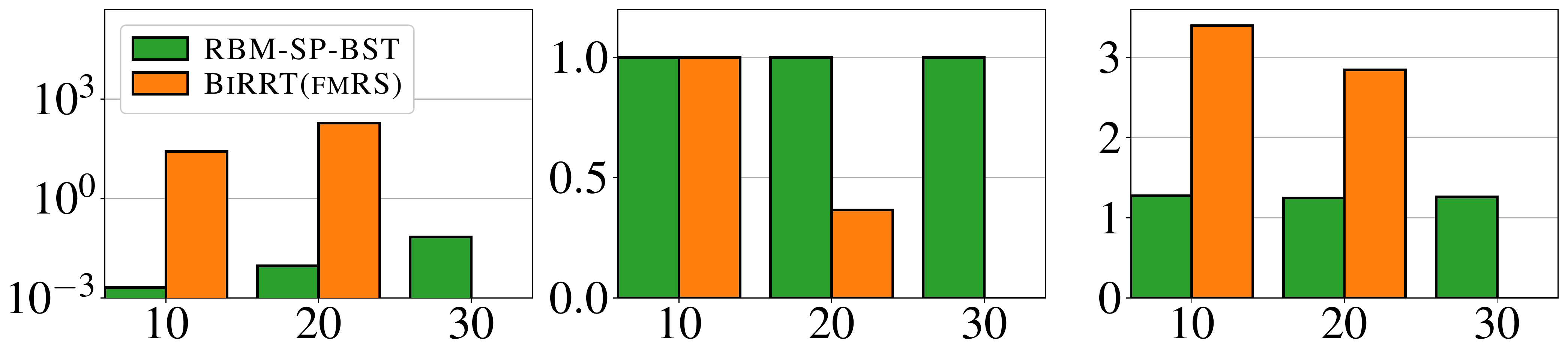}
    \caption{Comparison between methods in cuboid instances with $10$-$30$ cuboids and $\rho=0.3$ (left: computation time in seconds; middle: success rate; right: number of actions as multiples of $|\mathcal O|$).}
    \label{fig:StickComaprison}
\end{figure}

\subsection{Hardware Demonstration}
\begin{figure}[h!]
    \centering
    \includegraphics[width=\columnwidth]{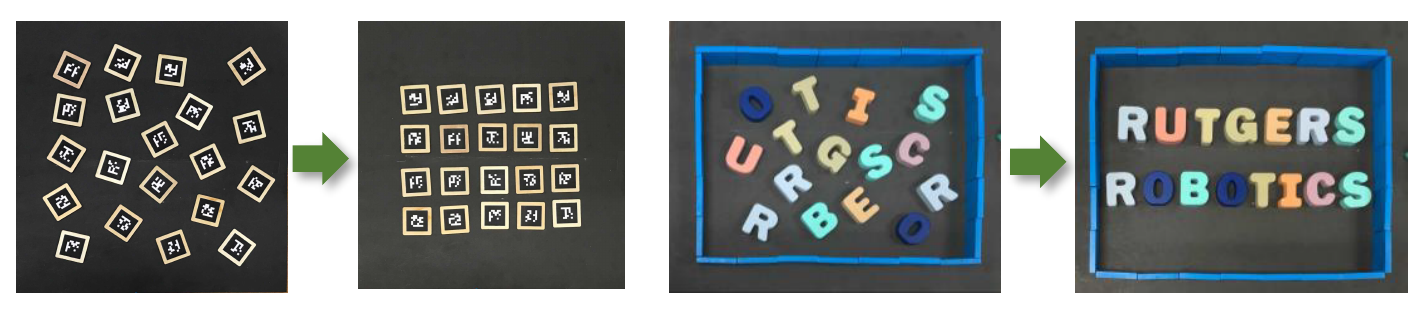}
    \caption{Experimental settings in addition to that of  Fig~\ref{fig:ApplicationExample}.}
    \label{fig:HardwareSetups}
\end{figure}

We further demonstrate that the plans computed by \trlb can be readily executed on real robots in a complete vision-planning-control pipeline. Our hardware setup consists of a UR-5e robot arm, an OnRobot VGC 10 vacuum gripper, and an Intel RealSense D435 RGB-D camera. As shown in the accompanying video, \trlb solves all attempted instances (Fig~\ref{fig:ApplicationExample} and Fig.~\ref{fig:HardwareSetups}), which involves concave objects, in an apparently natural and efficient manner. 
%

\section{Conclusion and Future Work}
The \trlb framework proposed in this work employs the dependency graph representation and a lazy buffer allocation approach for efficiently solving the problem of rearranging many tightly packed objects on a tabletop using internal buffers (\toroi). Extensive simulation studies show that \trlb  computes rearrangement plans of comparable or better quality as the state-of-the-art methods, and does so up to $2$ magnitudes faster. 
Demonstration with real robot experiments shows that the solutions computed by \trlb are not only efficient but also appear natural. As such, the solutions can potentially be deployed as part of home automation and industrial automation solutions in next generation robots. 



\bibliographystyle{IEEEtran}
\bibliography{c}

\end{document}